\documentclass[letterpaper, 10 pt, conference]{ieeeconf}

\usepackage{cite}

\usepackage[pdftex]{graphicx}
\usepackage{subcaption}
\usepackage{multirow}
\usepackage{makecell}
\usepackage{setspace}

\usepackage{pgfplots}
\pgfplotsset{width=7cm,compat=newest}

\usepackage{booktabs}
\usepackage{multirow}
\usepackage[hyphens]{url}
\usepackage{amssymb}
\usepackage{xcolor}
\let\labelindent\relax
\usepackage{enumitem}

\usepackage{amsmath}
\usepackage{wasysym}
\usepackage{units}
\usepackage[nolist]{acronym}
\usepackage{array}
\usepackage{balance}
\usepackage{cleveref}
\usepackage{textcomp}
\usepackage{pgfplotstable}
\usetikzlibrary{calc}
\usetikzlibrary{patterns}
\usepackage{units}
\usepackage{siunitx}
\pgfplotsset{compat=newest}

\pgfplotsset{
	percentage series/.style={
		table/y expr=(\thisrow{#1}),
		table/meta=#1,
		table/x expr=\coordindex,
	},
	percentage plot/.style={
		axis line style={draw=none},
		tick pos=lower,
		tick align=outside,
		tick style={draw=none},
		ymin=0,ymax=4,
		ytick={0,1,2,3,4},
		yticklabel=\pgfmathprintnumber{\tick}\,\unit{s},
		ymajorgrids=true,
		grid style={gray},
	}
}

\pgfplotstableread[col sep=comma,header=false]{
	,2.848471,0.
	,0.,3.934877
	,0.955633,0.
	,0.,0.876669
}\data
\pgfplotstablecreatecol[
create col/expr={\thisrow{1} + \thisrow{2}}
]{sum}{\data}

\definecolor{high_contrast}{RGB}{249, 223, 116}
\newcommand{\disable}[1]{}

\newcommand{\mycirc}[1][black]{\Large\textcolor{#1}{\ensuremath\bullet}}

\definecolor{dai_ligth_grey}{RGB}{230,230,230}
\definecolor{dai_ligth_grey20K}{RGB}{200,200,200}
\definecolor{dai_ligth_grey40K}{RGB}{158,158,158}
\definecolor{dai_ligth_grey60K}{RGB}{112,112,112}
\definecolor{dai_ligth_grey80K}{RGB}{68,68,68}

\definecolor{dai_petrol}{RGB}{0,103,127}
\definecolor{dai_petrol20K}{RGB}{0,86,106}
\definecolor{dai_petrol40K}{RGB}{0,67,85}
\definecolor{dai_petrol80}{RGB}{0,122,147}
\definecolor{dai_petrol60}{RGB}{80,151,171}
\definecolor{dai_petrol40}{RGB}{121,174,191}
\definecolor{dai_petrol20}{RGB}{166,202,216}

\definecolor{dai_deepred}{RGB}{113,24,12}
\definecolor{dai_deepred20K}{RGB}{90,19,10}
\definecolor{dai_deepred40K}{RGB}{68,14,7}

\definecolor{apfelgruen}{RGB}{140, 198, 62}
\definecolor{orange}{RGB}{244, 111, 33}
\definecolor{anthrazit}{RGB}{19, 31, 31}

\IEEEoverridecommandlockouts
\begin{document}
\title{\LARGE \bf A Benchmark for Spray from Nearby Cutting Vehicles}

\author{Stefanie Walz$^{1}$, Mario Bijelic$^{1,2}$, Florian Kraus$^{1}$, Werner Ritter$^{1}$, Martin Simon$^{3}$, Igor Doric$^{3}$
	\thanks{The authors are with $^{1}$Mercedes-Benz AG, $^{2}$Princeton University and $^{3}$MESSRING Active Safety GmbH.}
}

\maketitle

\thispagestyle{empty}
\pagestyle{empty}

\begin{acronym}
 \acro{CMOS}{complementary metal-oxide semiconductor}
 \acro{EU}{European Union}
 \acro{DENSE}{aDverse wEather eNvironment Sensing systEm}
 \acro{FIR}{far infrared} 
 \acro{NIR}{near infrared}
 \acro{SWIR}{short wave infrared}
 \acro{ADAS}{automotive drive assistance system}
 \acro{RMS}{root mean squared}
 \acro{LIDAR}[LiDAR]{light detection and ranging}
 \acro{RADAR}[RaDAR]{radio detection and ranging}
 \acro{TOF}{time of flight}
 \acro{NFIR}{near field infrared}
 \acro{FOV}[FoV]{field of view}
 \acro{OPAL}{obscurant penetrating autosynchronous lidar}
 \acro{InGaAs}{indium gallium arsenide}
 \acro{FMCW}{frequency-modulated continuous wave}
 \acro{IMU}{inertial measurement unit}
 \acro{AP}{average precision}
 \acro{IOU}[IoU]{intersection over union}
\end{acronym}
\begin{abstract}
Current driver assistance systems and autonomous driving stacks are limited to well-defined environment conditions and geo fenced areas.
To increase driving safety in adverse weather conditions, broadening the application spectrum of autonomous driving and driver assistance systems is necessary.
In order to enable this development, reproducible benchmarking methods are required to quantify the expected distortions.
In this publication, a testing methodology for disturbances from spray is presented.
It introduces a novel lightweight and configurable spray setup alongside an evaluation scheme to assess the disturbances caused by spray.
The analysis covers an automotive RGB camera and two different LiDAR systems, as well as downstream detection algorithms based on YOLOv3 and PV-RCNN.
In a common scenario of a closely cutting vehicle, it is visible that the distortions are severely affecting the perception stack up to four seconds showing the necessity of benchmarking the influences of spray.
\end{abstract}

\section{Introduction}
\begin{figure}[t!]
	\centering
	\hspace*{-3mm}
	\resizebox{1.05\columnwidth}{!}{
		\begin{tikzpicture}
		
		\node[anchor=west,inner sep=0pt] (clear_cam) at (0,0) {\includegraphics[height=20mm]{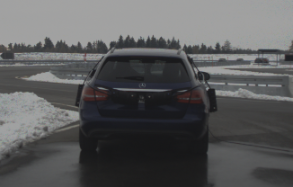}};
		\node[white] at ($(clear_cam) + (-0.9,-0.87)$) {\footnotesize RGB-Clear};
		
		\node[anchor=west,inner sep=0pt] (spray_cam) at (0,2.04) {\includegraphics[height=20mm]{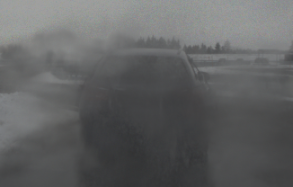}};
		\node[white] at ($(spray_cam) + (-0.9,-0.87)$) {\footnotesize RGB-Spray};
		
		\node[anchor=west,inner sep=0pt] (clear_lidar) at (3.18,0) {\includegraphics[height=20mm]{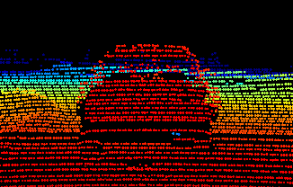}};
		\node[white] at ($(clear_lidar) + (-0.77,-0.87)$) {\footnotesize \acs{LIDAR}-Clear};
		
		\node[anchor=west,inner sep=0pt] (spray_lidar) at (3.18,2.04) {\includegraphics[height=20mm]{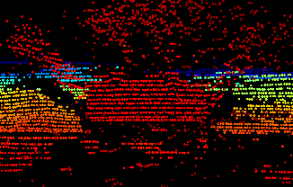}};
		\node[white] at ($(spray_lidar) + (-0.77,-0.87)$) {\footnotesize \acs{LIDAR}-Spray};
		
		\node[anchor=west,inner sep=0pt] (colorbar) at (3.11,-1.13) {\includegraphics[height=4.6mm]{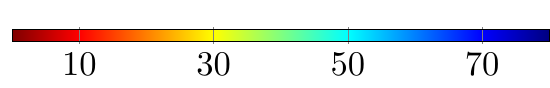}};
		\node at ($(colorbar) + (1.63,-0.02)$) {\tiny [m]};

	\end{tikzpicture}
}
\vspace*{-7mm}
	\caption{Influence of a spray plume generated by the proposed spray machine on camera and \acs{LIDAR} measurements (top) with clear reference on the bottom. }
	\label{fig:vehicle_in_spray_cloud}
	\vspace*{-6mm}
\end{figure}
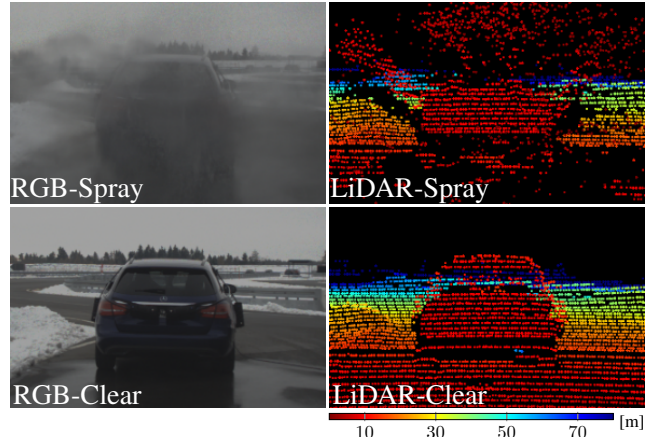

Safe autonomous driving requires a variety of sensors to obtain a detailed and redundant perception of the environment. Therefore, most systems cover more than two sensing modalities to achieve ASIL D safety certification~\cite{ASIL}. Currently, for a detailed perception of the vehicle's surroundings cameras, \ac{LIDAR}, and \ac{RADAR} systems are used as in \cite{caesar2019nuscenes,Bijelic_2020_STF}.
While all sensors perform reasonably in good weather, degraded conditions affect the perception in multiple ways.
Complications may arise from airborne particles, foreign contaminants of other road users, and self-soiling from mist and dirt particles on the ground according to \cite{Hagemeier2011}.
Until now, it has been shown that airborne particles from falling snowflakes~\cite{Michaud2015}, rain~\cite{hasirlioglu2016test}, and fog~\cite{bijelic2018benchmarking,bijelic2018benchmark,hasirlioglu2017reproducible} can degenerate the performance of single sensors. 
Another source of adversity is whirled up spray from a leading vehicle.
This influences the perception of autonomous driving systems as shown in Fig.~\ref{fig:vehicle_in_spray_cloud} and is currently only weakly investigated.
Since spray only requires a wet road, it occurs more frequently than heavy rainfall (5 times a year \cite{Myhre2019}), dense fog (12 times a year \cite{VanOldenborgh2010a}), or blizzards (13 times a year \cite{Coleman2017}). Specifically, this condition is fulfilled with a rainfall rate higher than \unit[1]{mm} and occurs about 100 times a year in northern America and Europe~\cite{DWD}. 
Despite the low rainfall rates, we are able to show that a challenging and common scenario of an overtaking and closely cutting vehicle in spray conditions can cause a total sensor blockage. 

In the past, experiments were made to investigate the spray of trucks \cite{shearman1998trials,spraytruck} and are summarized by the NHTSA in \cite{NHTSASprayStats}.
Shearman et al. \cite{shearman1998trials} present the only systematic investigation in the degeneration of perception capabilities and show that spray leads to a reduction of up to \SI{20}{dB} for \ac{LIDAR} signals caused by a truck at a speed of \SI{60}{\km\per\hour}.
Despite the sparsity of literature, the severity from road spray has been recognized, leading to the EEC Directive (91/226/EEC) enacted in 1991 mandating the installation of spray reduction devices for heavy vehicles.
However, deeper investigations are missing, as research was mainly centered on soiling from leading vehicles or the vehicle itself.
The investigations aimed, for example, to increase the clear view through the windshield or side windows \cite{spruss2016beitrag,Gaylard2017,kabanovs2019investigation, maycock1966problem, gaylard2011simulation} by aerodynamic optimizations. 

Within this work, we present a benchmarking approach to test reproducible spray, independent of ambient weather conditions.
Our approach consists of a portable spray machine attachable to a variety of vehicles and body shapes while simultaneously being independent of large infrastructures required for street-wettening systems.

During the experiments we found that:
\begin{itemize}
	\item A blockage time of up to four seconds can be expected for camera sensors during overtaking maneuvers.
	\item Spray degrades RGB images especially at night and with additional light sources as active brake lights.
	\item A \ac{LIDAR} with a spinning housing shows a \SI{22}{\m} long clutter cloud behind the spray vehicle at \SI{100}{\km\per\hour}.
	\item Active cleaning systems are essential for \ac{LIDAR} systems in spray conditions.
	\item Object detection algorithms for \ac{LIDAR} and camera suffer from false positives, inaccurate object dimensions, and missing detections in spray conditions.
	\item Qualitatively \ac{RADAR} systems are not affected by spray.	
\end{itemize}

\section{Related Work}
Vehicles driving over wet surfaces swirl up water from the road.
The adhesion of the tires causes water particles to be trapped in the tread profile.
When moving forward, the tires lift off the ground, and particles are detached in small water droplets, forming a spray plume behind the vehicle~\cite{schmiedel2019road}.
The size and the number of droplets depend mainly on the vehicle speed, road wetness, and tires. In general, the droplet size reduces with increasing rotation speed \cite{spruss2016beitrag, frey2005wse,Gaylard2017}. 
Previous research mainly focuses on the surface contamination of the vehicle caused by the spray \cite{kabanovs2019investigation, maycock1966problem, gaylard2011simulation}.
These studies distinguish between contamination generated by the own vehicle (self-soiling) and contamination caused by foreign influences (third-part soiling). Aerodynamic adjustments are applied to vehicle components to improve the vehicle's dirt resistance and the driver's visibility.
Investigations to analyze how tire spray can be used to determine the wetness of the road are carried out by \cite{schmiedel2019road, frey2005wse}.
Others show that through pneumatic mudflaps \cite{Goetz1995} the effects of spray can be minimized.
Unfortunately, only a small amount of work investigates the impact of road spray on automotive sensors~\cite{shearman1998trials}.

\subsection{Sensors in Adverse Environmental Conditions}
In recent years, a wide range of works have investigated the influence of adverse environmental conditions on automotive sensors.
Hasirlioglu \textit{et al.} \cite{hasirlioglu2016modeling} describe a theoretical model to determine the influence of fog and rain on camera, \ac{LIDAR}, and \ac{RADAR} systems.
To verify the model's theory, appropriate rain \cite{hasirlioglu2016test} and fog \cite{hasirlioglu2017reproducible} simulators were developed.
The evaluation shows that \ac{LIDAR} sensors are more affected by adverse weather conditions than \ac{RADAR} systems.
Moreover, adverse weather causes a decrease in contrast for the camera system.
Several kinds of research have indicated that this decay of contrast is caused by atmospheric particles scattering the light beams in various directions \cite{bijelic2018benchmarking,oakley1998improving, schechner2001instant, narasimhan2001removing}.
Many works seek to improve the degraded images by applying physical models \cite{oakley1998improving}, polarization \cite{schechner2001instant}, or normalized radiance calculations \cite{narasimhan2001removing}.
However, the performance of these algorithms is often quite subjective.
Peynot \textit{et al.}~\cite{peynot2009towards} has recorded a dataset in challenging environmental conditions (dust, smoke, rain) with an advanced perception system for autonomous driving vehicles.
The acquired data revealed that challenging environments cause a significant performance drop of \ac{LIDAR} systems since airborne dust particles are detected by laser sensors and hide obstacles behind the dust cloud.
This backscattering effect is also observed by Bijelic \textit{et al.} \cite{bijelic2018benchmark} in foggy conditions.
Compared to \ac{LIDAR} systems, the influence of backscatter onto automotive \ac{RADAR} sensors with a frequency of \SI{77}{\giga\hertz} is almost negligible \cite{brooker2007seeing}.

Due to a lack of information about the correlation of swirled-up road water and conventional rain, Shearman \textit{et al.} \cite{shearman1998trials} investigate the sensor behavior of \ac{LIDAR} and \ac{RADAR} systems in different spray conditions caused by test vehicles traveling over a wetted roadway.
Recordings were acquired by static sensors at the roadside and by sensors-equipped vehicles following the spray plume.
The results show that \ac{LIDAR} sensors suffer more from attenuation and clutter effects than \ac{RADAR} systems.
Compared to \cite{shearman1998trials}, this work focuses on the evaluation of camera and \ac{LIDAR} systems, as many investigations have shown that \ac{RADAR} systems are less affected by adverse conditions.

\subsection{Spray Generation Methods}
There exist multiple methods to simulate swirled-up water from the road.
In \cite{potthoff1974untersuchung,widdecke2001moderne, kuthada2002advanced} spray plumes are generated in wind tunnels using stationary installed spray systems.
The advantage of this method is that reproducible spray patterns are generated by controlling and adapting the nozzles.
In contrast, dynamic driving tests on wetted roads \cite{borg2004development,shearman1998trials} have problems with reproducibility since external weather factors are hard to control, and maintaining a constant water quantity on the road surface is a challenging task.
However, real driving tests consider the influence of moving car dynamics and provide extremely realistic spray patterns.

To generate spray in a wind tunnel, Potthoff \cite{potthoff1974untersuchung} introduces a spray bar with up to 16 nozzle sets that can be adjusted to the vehicle width.
One nozzle set consists of a water nozzle and a perpendicular compressed air nozzle to nebulize the emitted water droplets.
By positioning a height-adjustable rack with several spray nozzles directly in front of a thermo-wind tunnel opening, Widdecke \textit{et al.} \cite{widdecke2001moderne} and Kuthada \textit{et al.} \cite{kuthada2002advanced} are able to produce volume flows of \SI{200}{\liter\per\hour} to a maximum of \SI{1200}{\liter\per\hour}.

For dynamic driving tests, Shearman \textit{et al.} \cite{shearman1998trials} use spray modules at the roadside to water the road constantly.
The undefined amount of road water causes variations in spray patterns and complicates reproducible tests.
To improve the consistency of the water quantity, a sprinkling system is developed in \cite{bebb2012road}, which incorporates several spray modules for the roadside alongside a road condition sensor to reduce variations.
However, watering the entire road for spray generation requires a large amount of water and leads to problems in freezing conditions.
To overcome this issue, Zlocki \textit{et al.} \cite{spraytruck} introduce a truck-mounted spray machine for dynamic tests consisting of six water tanks and a pump.
For spray pattern variations, five separate switchable circles and different nozzle configurations are provided.
A vehicle-mounted spray machine allows to incorporate moving car dynamics and to produce reproducible spray patterns simultaneously.
In contrast to the spray truck, our proposed spray machine has the advantage of not being limited to one vehicle but rather can be mounted on any available vehicle.

\begin{table}[t]
	\vspace*{1.7mm}
	\centering
	\caption{Technical summary of the tested sensors. Differences between similar models are marked \textcolor{red}{red}.}
	\vspace*{-1mm}
	\resizebox{\linewidth}{!}{
		\begin{tabular}{@{}clll@{}}
			\toprule
			\textbf{Sensor Property} & \textbf{Stereo Camera} & \textbf{Radar} & \textbf{Lidar} \\
			\midrule
			\textbf{Make} & OnSemi & - & Velodyne \\
			\textbf{Type} & AR0230AT &  Proprietary & HDL64S3D/\textcolor{red}{VLP32C}\\
			\textbf{Sensor Count} & 2 & 1 & 2\\
			\textbf{Quantization} & \SI{12}{bit} & - & \SI{16}{bit}\\
			\textbf{Framerate} & \SI{30}{\hertz} & \SI{15}{\hertz} & \SIrange{5}{20}{\hertz}\\
			\textbf{Timeync} & Upon Arrival & Upon Arrival & GPS\\
			\textbf{FoV} & \SI{39.6}{\degree}$\times$\SI{21.6}{\degree} & \SI{17}{\degree}/\SI{56}{\degree} (far/close) &  $\SI{360}{\degree}\times\SI{26.9}{\degree}$/\textcolor{red}{\SI{40}{\degree}}\\
			\textbf{Wavelength} & \SIrange{380}{740}{\nm} & \SI{4}{\mm} & \SI{903}{\nm}\\
			\textbf{Angular Resolution} & \SI{0.020}{\degree} &\SI{1}{\degree}/\SI{4}{\degree} (far/close)  & \SIrange{0.0864}{0.3}{\degree}\\
			\textbf{Illumination} & Passive & Active & Active \\
			\bottomrule
	\end{tabular}}
	\label{tab:sensor_specs}
	\vspace*{-2mm}
\end{table}

\begin{figure}[t]
	\includegraphics[width=0.94\columnwidth]{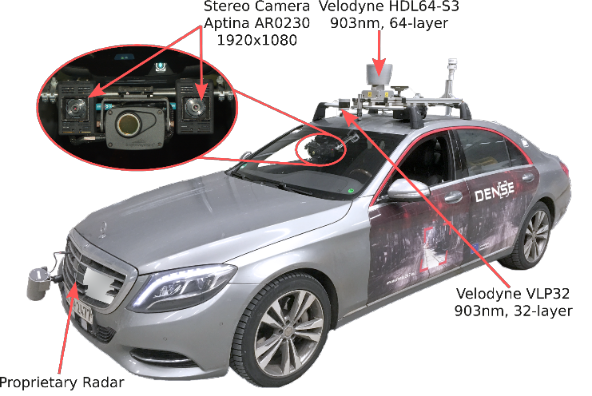}
	\vspace*{-1mm}
	\caption{Sensor setup of the test vehicle.}
	\label{fig:sensor_setup}
	\vspace*{-5.5mm}
\end{figure}

\section{Tested Sensor Setup}
To evaluate the performance of automotive sensors in spray conditions, we equipped a test vehicle with a standard RGB stereo camera (OnSemi Aptina AR0230) and two state-of-the-art \ac{LIDAR} systems (Velodyne HDL64-S3, Velodyne VLP32C).
Both laser scanners are mounted on the car roof, while the camera is located behind the windshield, which is illustrated in Fig.~\ref{fig:sensor_setup}.
The stereo camera runs at \SI{30}{\hertz} and offers a resolution of $1920\times 1024$ pixels with 12-bit quantization.
The two laser scanners from Velodyne are operating at \SI{903}{\nm} and provide dual returns (strongest and last) at \SIrange{5}{20}{\hertz}.
The Velodyne HDL64-S3D provides 64 equally distributed scanning lines and a mechanism for spinning the entire sensor unit vertically, which delivers a \ac{FOV} of $\SI{360}{\degree}\times\SI{26.9}{\degree}$ and ranges up to \SI{120}{\m}.
In contrast, the Velodyne VLP32C offers 32 non-linear distributed scanning lines and an internal spinning mechanism to support a \ac{FOV} of $\SI{360}{\degree}\times \SI{40}{\degree}$ and ranges up to \SI{200}{\m}.
Additionally, a proprietary \ac{FMCW} \ac{RADAR} system is used to measure the distance and speed of the leading spray vehicle.
The \ac{RADAR} operates at a frequency of \SI{77}{\giga\hertz} and is located in the radiator grille of the testing vehicle.
A technical summary of all sensors can be found in Table~\ref{tab:sensor_specs}.

\section{Spray Test Machine}
The \emph{Spray Test Machine} \cite{spray2019patent, spray2021datasheet} was developed in order to enable repeatable tests in spray conditions independent of the current local weather.
The machine creates a reproducible spray corridor behind a vehicle (Fig.~\ref{fig:experiment_setup}a) which allows an evaluation of the sensor performance of a trailing vehicle in challenging but reproducible weather conditions.
All required components are installed in a leading spray vehicle.

\begin{figure}
\vspace*{1.3mm}
\includegraphics[width=0.99\columnwidth]{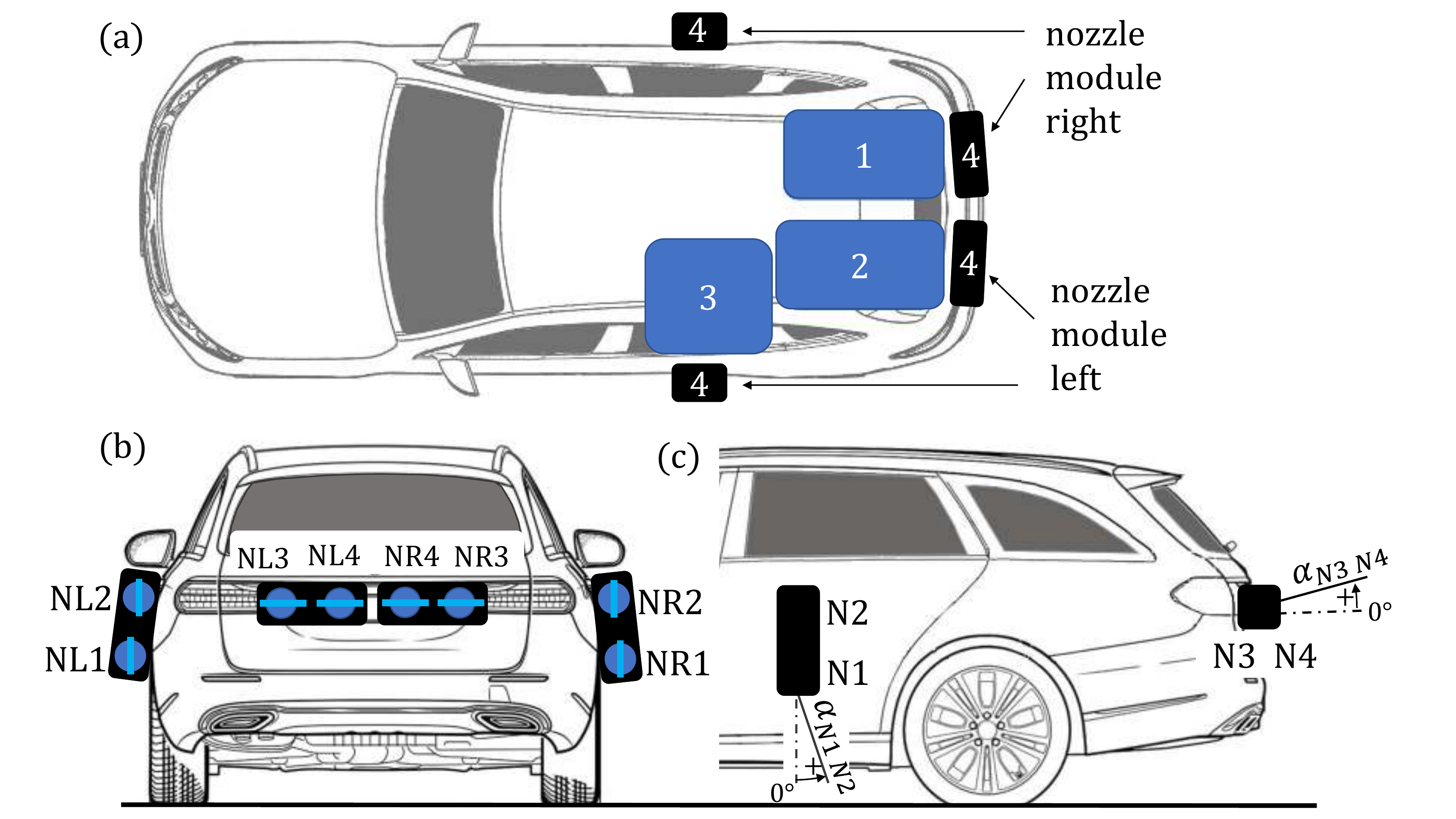}
\caption{Overview of the technical spray setup: (a)
	1:~high-performance pump module;
	2:~electric power module;
	3:~\SI{100}{\liter} mobile water tank;
	4:~spray modules with two spray nozzles each.
(b) Mounting positions of the spray nozzles. (c) Side view and spray nozzle adjustment.}
\label{fig:setupSpray}
\vspace*{-4mm}
\end{figure}

The setup consists of four modular components:
A high-performance pump, an electric power module, a mobile water tank (\SI{100}{\liter}), and four spray modules, each equipped with two flat jet spray nozzles. A illustration is shown in Fig.~\ref{fig:setupSpray}a.

The installation of the spray setup does not require any modifications to any of the involved vehicles.
The spray modules are made of flexible material and adapt to the shape of the vehicle.
In our experiments, the spray modules are attached to the back door (close to the rear wheels) and to the trunk lid with tensioning straps, cf. Fig.~\ref{fig:setupSpray}b and~3c.

The spray direction of each spray module and each individual nozzle can be variably adjusted as illustrated in Fig.~\ref{fig:setupSpray}c.
Furthermore, different types of nozzles can be mounted to the spray modules.
The size of the spray droplets and the droplet density within the spray corridor can be varied by the type of nozzle.
These options enable different spray patterns and thus enable various reproducible test conditions at the proving ground.

\begin{figure}
\includegraphics[width=0.99\columnwidth]{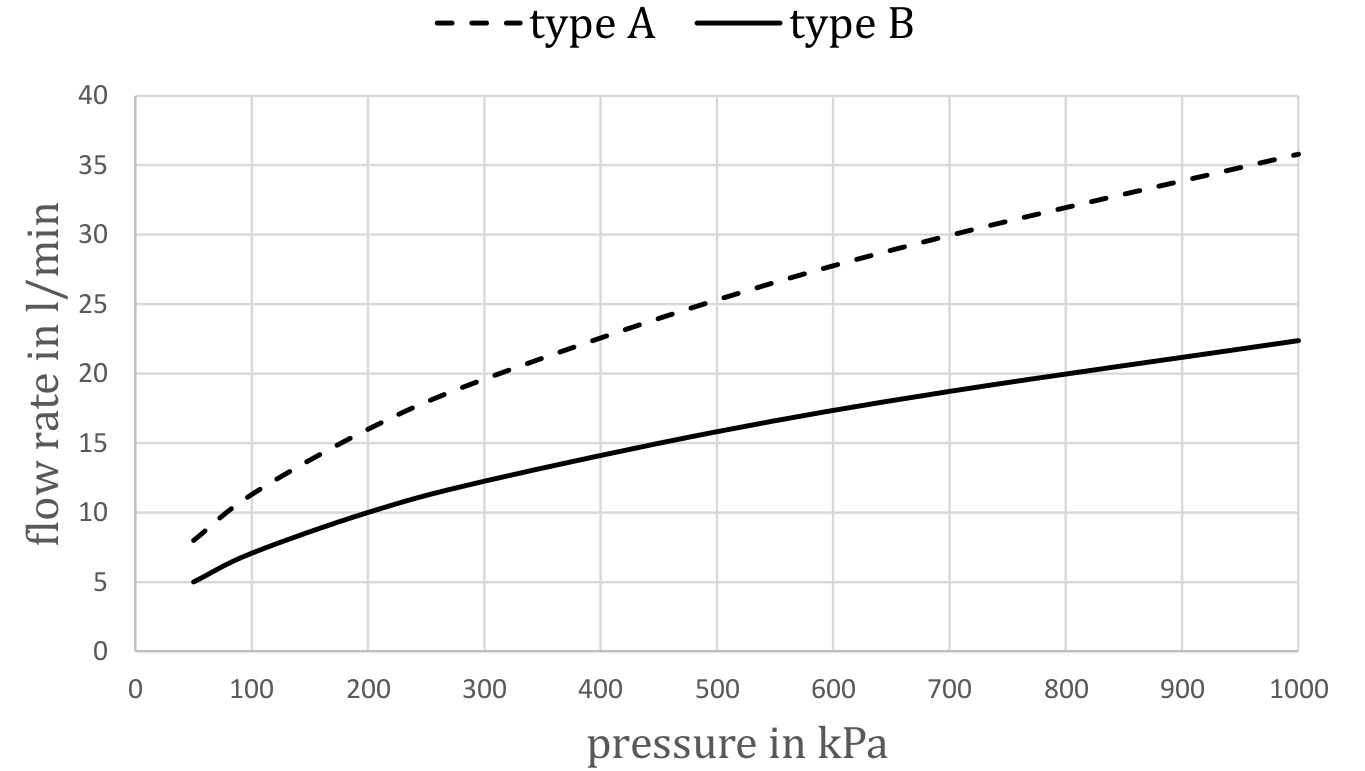}
\caption{Characteristics of flat jet nozzle type A (\SI{20}{\degree} opening) and type B (\SI{90}{\degree} opening).}
\label{fig:nozzle}
\vspace*{-6mm}
\end{figure}

During the experiments, two types of nozzles with different flow rates are used.
Detailed characteristics of the flow rates are illustrated in Fig.~\ref{fig:nozzle}.
The nozzle mounting positions within the modules are shown in Fig.~\ref{fig:setupSpray}b.
To obtain realistic spray patterns, the nozzle configurations of Table~\ref{tab:nozzle-config} are applied.
Different settings are used for static and dynamic tests.
The missing airstream for a stationary spray vehicle is compensated by higher water pressure.
\begin{table}
	\vspace*{1.7mm}
	\renewcommand{\arraystretch}{0.95}
	\centering
	\caption{Nozzle setup for dynamic and static tests. Type~A: high flow rate; Type~B: low flow rate. Mounting positions: Fig.~\ref{fig:setupSpray}b.}\label{tab:nozzle-config}
	\resizebox{\columnwidth}{!}{
	\begin{tabular}{ccc}
		\toprule
		\textbf{Mounting Position} & \textbf{Dynamic Tests} & \textbf{Static Tests} \\ \midrule
		    NL1 / NR1     &    Type A    &  No Nozzle  \\
		    NL2 / NR2     &  No Nozzle   &   Type A / Type B   \\
		    NL3 / NR3     &    Type B    &   Type A / Type B   \\
		    NL4 / NR4     &    Type B    &  No Nozzle  \\ \bottomrule
	\end{tabular}}
\vspace*{-5mm}
\end{table}

\section{Experimental Setup}
\begin{figure}
 \vspace*{1.7mm}
 \centering
 \includegraphics[width=0.95\columnwidth]{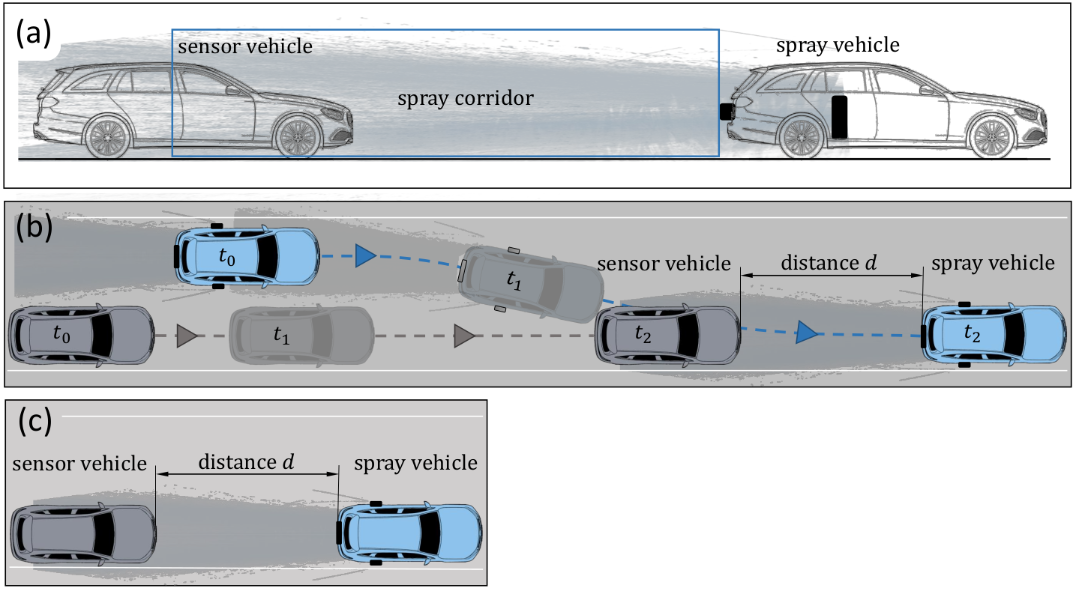}
 \caption{Experimental test setup. (a) Definition of the spray corridor; (b) Dynamic test setup, the spray vehicle (blue) overtakes the test vehicle (gray) and cuts in front. As a result the test vehicle is affected by the spray corridor; (c) Static test setup, spray vehicle is placed in front of the test vehicle to allow for safe experiments without windshield cleaning.}
 \label{fig:experiment_setup}
 \vspace*{-5mm}
\end{figure}

Well-defined and repeatable weather conditions are essential for testing and comparing different sensors.
To assess the effects of a spray plume on the sensor setup, we split the experiments into dynamic and static tests.
The dynamic tests consist of an overtaking maneuver where the spray vehicle cuts in front of the test vehicle equipped with the sensor setup.
This is illustrated in Fig.~\ref{fig:experiment_setup}b.
During the tests, the spray system is continuously activated, resulting in a defined spray corridor, cf. Fig.~\ref{fig:experiment_setup}a.

The dynamic tests are carried out at two different speed configurations.
First, the test vehicle drives at \SI{60}{\km\per\hour}, and the spray vehicle overtakes at \SI{70}{\km\per\hour}.
In the second setup, the test vehicle has a speed of \SI{80}{\km\per\hour}, and the overtaking spray vehicle drives at \SI{100}{\km\per\hour}.
All dynamic tests are carried out with the windshield wipers set to automatic mode and maximum speed.
Dynamic tests without cleaning the windshield are not feasible due to obvious safety issues.

To conduct experiments without active window cleaning, static tests are performed.
The spray vehicle is positioned \SI{10}{\meter} in front of the test vehicle to simulate the critical point of an overtaking maneuver, cf. Fig.~\ref{fig:experiment_setup}.
This distance ensures that the sensor setup is completely contained in the spray corridor, simulating a worst-case scenario.
In this way, our spray setup allows for a safe evaluation of spray without using windshield wipers.
All tests, dynamic and static, are conducted during day and night settings.
The static tests are performed with and without active windshield cleaning.

To assess the effects of the spray on object detection algorithms, we train deep learning based object detection neural networks for \ac{LIDAR} and camera data.
For the \ac{LIDAR} system, a PointVoxel-RCNN (PV-RCNN) \cite{Shi_2020_CVPR} architecture is trained on the dataset provided by \cite{Bijelic_2020_STF}, which was collected by the same test vehicle as used in this work.
PV-RCNN is a two-stage approach that first encodes the point cloud into a voxel grid which is then processed by sparse 3D convolutions.
The 3D convolutional network predicts 3D proposals, which are consecutively refined and classified in a second stage.
The final outputs of this network are 3D bounding boxes with a class prediction.
For the camera data, a YOLOv3 \cite{Redmon2018} object detection network is trained on the image data provided by the same dataset \cite{Bijelic_2020_STF}.
YOLOv3 is a fully convolutional neural network architecture that detects 2D bounding boxes of road users.

\section{Evaluation}
\begin{figure*}[t]
\vspace*{1.7mm}
\centering 
\resizebox{\textwidth}{!}{
\begin{tabular}{>{\centering\arraybackslash}m{0.2cm} >{\centering\arraybackslash}m{0.2cm} 
>{\centering\arraybackslash}m{0.24\textwidth} >{\centering\arraybackslash}m{ 
0.24\textwidth} >{\centering\arraybackslash}m{0.24\textwidth} >{\centering\arraybackslash}m{0.24\textwidth}}
& &
\multicolumn{2}{c}{RGB Image}
 & 
\multirow{2}{*}{\makecell[c]{HDL64-S3D + \\ \acs{RADAR} \mycirc[green]}}& 
\multirow{2}{*}{\makecell[c]{VLP32C + \\ \acs{RADAR} \mycirc[green]}} \\
& &
day & night
&  & \\
\rotatebox[origin=l]{90}{reference} & &
\includegraphics[width=0.25\textwidth]{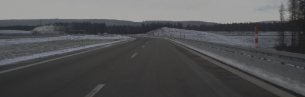} 
& 
\includegraphics[width=0.25\textwidth]{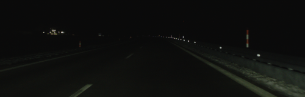} 
& 
\includegraphics[width=0.25\textwidth]{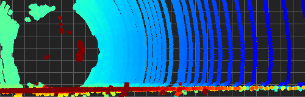} 
&
\includegraphics[width=0.25\textwidth]{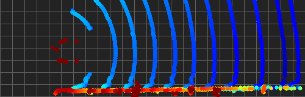} 
\\ 

\multirow{2}{*}{\rotatebox{90}{\parbox[c]{1.8cm}{\centering $d = \SI{13}{\m}$}}} &
\rotatebox{90}{\footnotesize \SI{70}{\km\per\hour}} &
\includegraphics[width=0.25\textwidth]{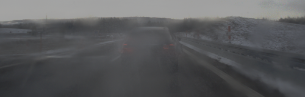} 
& 
\includegraphics[width=0.25\textwidth]{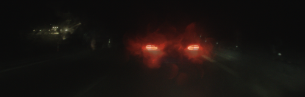} 
& 
\includegraphics[width=0.25\textwidth]{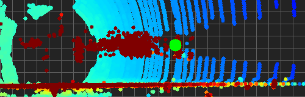} 
&
\includegraphics[width=0.25\textwidth]{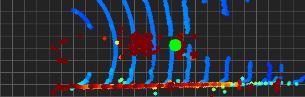} 
\\

&
\rotatebox{90}{\footnotesize \SI{100}{\km\per\hour}} &
\includegraphics[width=0.25\textwidth]{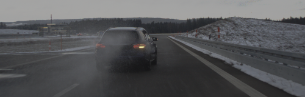} 
& 
\includegraphics[width=0.25\textwidth]{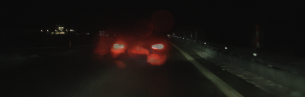} 
& 
\includegraphics[width=0.25\textwidth]{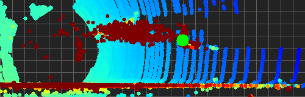} 
&
\includegraphics[width=0.25\textwidth]{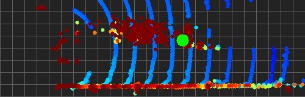} 
\\

\multirow{2}{*}{\rotatebox{90}{\parbox[c]{1.8cm}{\centering $d = \SI{17}{\m}$}}} &
\rotatebox{90}{\footnotesize \SI{70}{\km\per\hour}} &
\includegraphics[width=0.25\textwidth]{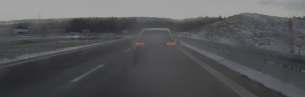} 
& 
\includegraphics[width=0.25\textwidth]{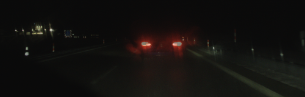} 
& 
\includegraphics[width=0.25\textwidth]{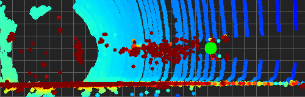} 
&
\includegraphics[width=0.25\textwidth]{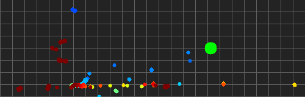} 
\\

&
\rotatebox{90}{\footnotesize \SI{100}{\km\per\hour}} &
\includegraphics[width=0.25\textwidth]{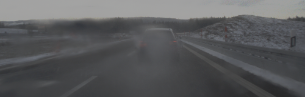} 
& 
\includegraphics[width=0.25\textwidth]{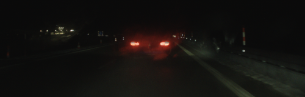} 
& 
\includegraphics[width=0.25\textwidth]{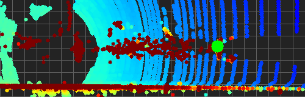} 
&
\includegraphics[width=0.25\textwidth]{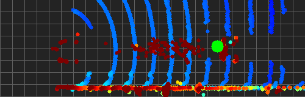} 
\\

\multirow{2}{*}{\rotatebox{90}{\parbox[c]{1.8cm}{\centering $d = \SI{20}{\m}$}}} &
\rotatebox{90}{\footnotesize \SI{70}{\km\per\hour}} &
\includegraphics[width=0.25\textwidth]{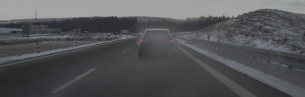} 
& 
\includegraphics[width=0.25\textwidth]{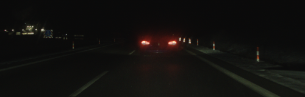} 
& 
\includegraphics[width=0.25\textwidth]{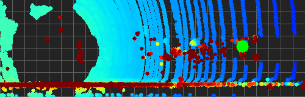}
&
\includegraphics[width=0.25\textwidth]{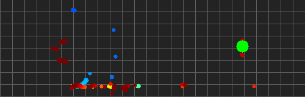} 
\\

&
\rotatebox{90}{\footnotesize \SI{100}{\km\per\hour}} &
\includegraphics[width=0.25\textwidth]{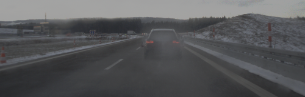} 
& 
\includegraphics[width=0.25\textwidth]{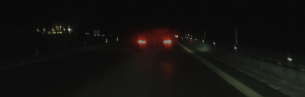} 
& 
\includegraphics[width=0.25\textwidth]{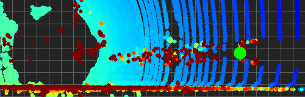} 
&
\includegraphics[width=0.25\textwidth]{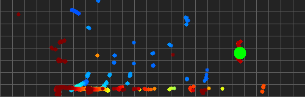} 
\\

\multirow{2}{*}{\rotatebox{90}{\parbox[c]{1.8cm}{\centering $d = \SI{24}{\m}$}}} &
\rotatebox{90}{\footnotesize \SI{70}{\km\per\hour}} &
\includegraphics[width=0.25\textwidth]{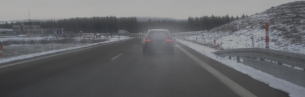} 
& 
\includegraphics[width=0.25\textwidth]{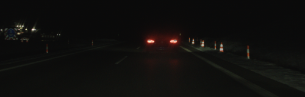}
& 
\includegraphics[width=0.25\textwidth]{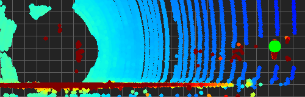}
&
\includegraphics[width=0.25\textwidth]{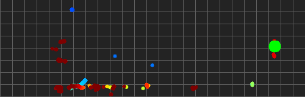} 
\\

&
\rotatebox{90}{\footnotesize \SI{100}{\km\per\hour}} &
\includegraphics[width=0.25\textwidth]{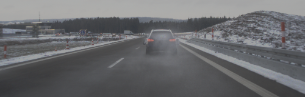} 
& 
\includegraphics[width=0.25\textwidth]{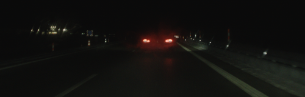} 
& 
\includegraphics[width=0.25\textwidth]{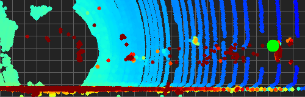} 
&
\includegraphics[width=0.25\textwidth]{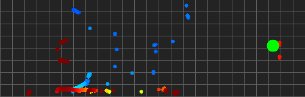} 
\\

 \end{tabular}}
 \vspace*{-3pt}
 \caption{RGB images of the camera and bird's-eye view of the laser scanners with overlayed radar measurements for different spray vehicle distances and speeds during overtaking. The color for the \acs{LIDAR} measurements encode the points height, and the  \textcolor{green}{green} dot indicates the \acs{RADAR} measurement of the spray vehicle. Since the Velodyne VLP32C does not contain a self-cleaning system, the sensor gets blocked by water droplets and the number of detected laser points decrease.}
 \label{fig:lidar_images_velodyne}
 \vspace*{-5mm}
\end{figure*}

\label{sec:eval}
\subsection{Qualitative Evaluation}
For a first impression of the sensor performance in spray, we qualitatively assess the sensor data of the laser scanners, the camera, and the \ac{RADAR} system.
Fig.~\ref{fig:lidar_images_velodyne} shows the RGB images of the camera during day and night as well as the bird's-eye views of the strongest return for the Velodyne HDL64-S3D and the VLP32C with overlayed \ac{RADAR} measurements. The colors of the \ac{LIDAR} points encode their height, and the green dot indicates the \ac{RADAR} measurement of the spray vehicle.
Measurements are acquired for different spray vehicle distances $d$ and speeds.
In general, road spray causes the impairment of the sensor performance based on two reasons. Firstly, by swirling water that adheres directly to the sensor or the windshield and immediately blocks the perception. 
Secondly, by particles in the air that act like small "lenses" and refract and absorb light rays.

The adhering water droplets lead to a blocked \ac{FOV} around a leading vehicle which occurs if the spray vehicle is in close proximity and the sensor setup is completely located inside the spray corridor as illustrated in Fig.~\ref{fig:experiment_setup}.
Especially sensors without (self-) cleaning systems are affected and get easily blocked, such as the Velodyne VLP32C.
However, the RGB images in Fig.~\ref{fig:lidar_images_velodyne} verify that even with an active cleaning system, the sensor view cannot be kept clear for close distances since spray permanently hits the windshield.
Only the Velodyne HDL64-S3D is not impacted by droplets on the sensor surface, since the rotation of the entire unit causes spray to be thrown away by centrifugal forces.

The second adversity, namely the airborne spray, creates a long tail behind a vehicle.
This weakens the contrast in the camera images locally around the spray vehicle and creates clutter behind the leading vehicle in the \ac{LIDAR} measurements.

Only the \ac{RADAR} sensor is not affected by spray conditions and is able to detect the spray vehicle ahead constantly, marked as a green dot in Fig.~\ref{fig:lidar_images_velodyne}.
Therefore, in this specific use case, the \ac{RADAR} is undisturbed, and we omitted the quantitative evaluation in the following.

\begin{figure}[t]
\vspace*{1mm}
\centering
\begin{minipage}[t]{\columnwidth}
\centering
\setlength{\tabcolsep}{0.5mm}
\resizebox{0.97\columnwidth}{!}{
\begin{tabular}{@{}ccc@{}}
\footnotesize Erroneous Detections & \footnotesize  Inaccurate Detections & \footnotesize  No Detections \\
\includegraphics[width=0.32\columnwidth]{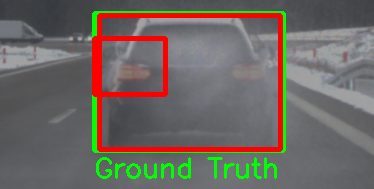} &
\includegraphics[width=0.32\columnwidth]{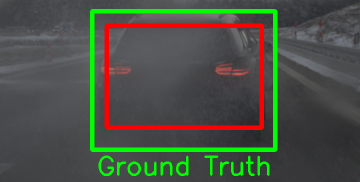} &
\includegraphics[width=0.32\columnwidth]{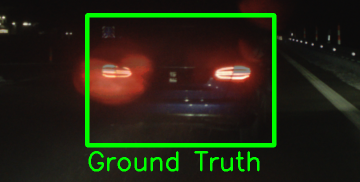}\\
\includegraphics[width=0.32\columnwidth]{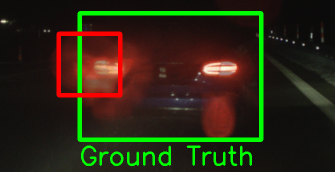} &
\includegraphics[width=0.32\columnwidth]{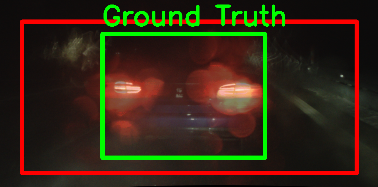} &
\includegraphics[width=0.32\columnwidth]{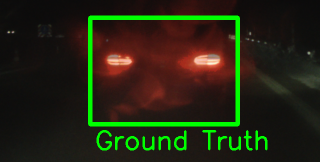}\\
\end{tabular}}
\vspace*{-2mm}
\subcaption{\small 2D-detection problems include false positives around the rear lights, wrong object dimensions, and missing detections.}
\label{fig:detection_problems_cam}
\end{minipage}

\begin{minipage}[t]{\columnwidth}
\centering
\vspace*{0mm}
\setlength{\tabcolsep}{0.5mm}
\resizebox{0.97\columnwidth}{!}{
\begin{tabular}{@{}cc@{}}
\footnotesize  Ghost detections (HDL64/VLP32) & \footnotesize  No detections/Blockage (VLP32) \\
\includegraphics[width=0.49\columnwidth]{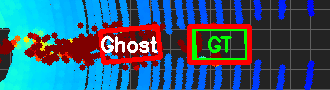} &
\includegraphics[width=0.49\columnwidth]{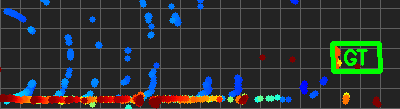}\\
\includegraphics[width=0.49\columnwidth]{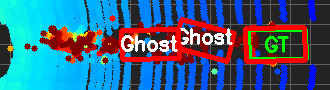} &
\includegraphics[width=0.49\columnwidth]{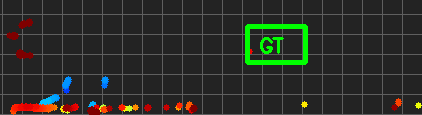}\\
\end{tabular}}
\subcaption{\small Spray causes 3D \ac{LIDAR} object detectors to hallucinated objects within the spray clutter and miss detections due to a blocked sensor.}
\label{fig:detection_problems_lidar}
\end{minipage}
\vspace*{-1mm}
\caption{Problems of detection algorithms caused by spray-induced decrease in contrast and clutter effects.}
\label{fig:detection_problems}
\vspace*{-6mm}
\end{figure}
\subsection{Qualitative Object Detection Performance in Spray}
We notice that our object detectors face several problems within the challenging spray conditions. First qualitative results are illustrated in Fig~\ref{fig:detection_problems}.
\subsubsection{Camera Based Detection}
Spray affects camera-based systems in two ways: First, scattering of light caused by the fine water droplets.
Second, water build-up on the windscreen, which can be reduced by windshield wipers.
Both result in a decrease of contrast within the recorded image data, cf. Fig.~\ref{fig:lidar_images_velodyne}.
This loss of contrast results in unclear edges for relevant objects, such as cars, causing multiple problems compared to clear weather scenarios.
First, this leads to worse localization, where the object dimensions are either predicted as too large or too small. Both are likely caused by missing information regarding the object dimensions.
Second, small false positive detections appear around the rear lights.
The detector confuses the rear lights with cars in the distance.
And lastly, even completely missing detections occur, which is most notable during night scenarios.
Examples of all scenarios are given in Fig.~\ref{fig:detection_problems_cam}.

\subsubsection{\ac{LIDAR} Based Detection}
Spray causes clutter within the \ac{LIDAR} point cloud, cf. Fig.~\ref{fig:lidar_images_velodyne}.
This strip of clutter is most notable for the HDL64-S3D sensor, which spins of adhering water droplets due to its rotation.

The clutter within the point cloud causes ghost detections where the detector "hallucinates" objects within the clutter.
Those hallucinated objects occur more frequently at higher speeds.
Furthermore, the object detector is unable to detect the spray vehicle with the sparse data from the blocked Velodyne VLP32C \ac{LIDAR} sensor, cf. Fig.~\ref{fig:detection_problems_lidar}.

\subsection{Quantitative Evaluation of the Dynamic Experiment}
For the quantitative evaluation of the dynamic experiments, the subjective blockage time of the camera and the length of the dragged spray tail are ascertained for different vehicle speeds. To measure the spray length, the longitudinal spread of the spray clutter is determined from the Velodyne HDL64-S3D data. Furthermore, the sensor measurements of the camera and the laser scanners are manually reviewed to evaluate how often the spray vehicle is detected by the object detection algorithms during overtaking.
\subsubsection{Camera Blockage Time}
Fig.~\ref{fig:camera_blockage} illustrates the blockage time of the camera for the different overtaking maneuvers at day and night. The blockage time is determined according to the subjective assessment of a human observer and his ability to steer the car safely, based on the RGB images. The results show that for a speed difference of \SI{10}{\km\per\hour} between test and spray vehicle, the camera is blocked for around \SIrange{3}{4}{\second}, while a speed difference of \SI{20}{\km\per\hour} causes a blockage of only \SI{1}{\second}. It should be noted that the sensory perception is constrained despite the active windshield wipers.
Due to different merging distances of the spray vehicle ahead, the blockage times vary between day and night, especially for low speed differences.

\subsubsection{Spray length}
The length of the spray tail dragged behind the vehicle increases for higher speeds, varying from \SI{13.5}{\m} at \SI{60}{\km\per\hour} to \SI{22}{\m} at \SI{100}{\km\per\hour} as illustrated in Fig.~\ref{fig:spray_length}.
Knowing the spray length at different vehicle speeds helps to understand at which distances the sensor setup can get blocked directly by water droplets and at which point only airborne water particles influence the sensor performance.
Thus, for the evaluation of the object detection algorithm, it is possible to distinguish between the time of the test vehicle being inside and the time of being outside the spray corridor.

\subsubsection{Object detection rate}

Fig.~\ref{fig:2d_detection_dynamic} and Fig.~\ref{fig:3d_detection_dynamic} show the performance of the 2D-camera and the 3D-\ac{LIDAR} detector, respectively, for different distances during the overtaking maneuver with \SI{70}{\km\per\hour}. The 2D detection results indicate that the algorithm has no trouble recognizing the spray vehicle in bright daylight. However, at night and distances between \SI{8}{\m} and \SI{16}{\m}, the detection performance decreases significantly. For these distances, the tested sensor setup is completely covered by the generated spray of the vehicle in front. Increasing the distance to the spray vehicle from \SI{8}{\m} to \SI{16}{\m} at a speed difference of \SI{10}{\km\per\hour} requires approximately three seconds. This equals the subjective determined blockage time of the camera for the corresponding vehicle speeds. When the spray vehicle moves on far enough and the test vehicle gets out of the spray corridor, the number of detection increases. This implies that water droplets on the windshield influence the camera sensor more than the water particles in the air. Additionally, the higher detection rate for increasing distances indicates that a spray cloud has less impact the further its distance from the perceiving sensor.

For the evaluation of the \ac{LIDAR} systems, the true detection rate of the spray vehicle is calculated in Fig.~\ref{fig:3d_detection_dynamic_vehicle}, and the false positive detection rate of the spray clutter is determined in Fig.~\ref{fig:3d_detection_dynamic_ghost}.
Fig.~\ref{fig:3d_detection_dynamic_vehicle} demonstrates that the 3D object detector constantly recognizes the spray vehicle in the point clouds of the Velodyne HDL64-S3D for all distances. However, the well visible spray plume in the Velodyne HDL64-S3D data (cf. Fig.~\ref{fig:lidar_images_velodyne}) causes the detection algorithm to mistake clusters from the clutter as actual objects leading to the high ghost detection rate in Fig.~\ref{fig:3d_detection_dynamic_ghost}.  
In contrast, the detection performance of the Velodyne VLP32C breaks down in Fig.~\ref{fig:3d_detection_dynamic_vehicle}. Since the Velodyne VLP32C is not equipped with a (self-) cleaning system, water droplets accumulate on the sensor surface resulting in a blockage of the sensor perception. 
However, the airstream of the test vehicle contributes to the self-cleaning process of the laser scanner and leads to a temporary increase in performance, where single points from the leading vehicle are captured up to a distance of \SI{28}{m}. From this distance onward the thin water film causes an attenuation which prevents capturing any point at further distances.
Due to the blockage of the sensor surface, the Velodyne VLP32C is not able to perceive clutter echoes caused by the spray plume, which results in a remarkably reduced number of ghost detections. 

For higher overtaking speeds, the effects are shifted to further distances and impair the performance longer, e.g. the ghost detection rate of 100\% can be measured from \SIrange{20}{32}{\m} instead of \SIrange{16}{24}{\m} due to the longer clutter plume.

\subsection{Quantitative Evaluation of the Static Experiment}
\begin{figure}
\vspace*{1mm}
\centering
\resizebox{0.95\columnwidth}{!}{
\begin{tikzpicture}[font=\Large]
\begin{axis}[
ybar stacked,
width=13cm,
height=5cm,
percentage plot,
bar width=0.9cm,
xmin=-.5,xmax=3.5,
xtick=data,
xticklabels from table={\data}{[index]0},
minor xtick={-.5,1.5,3.5},
minor tick length=1.4cm,
minor x tick style={draw,/pgfplots/every axis grid},
extra x ticks={0.5,2.5},
extra x tick labels={Test vehicle \SI{60}{\km\per\hour} \\ Spray vehicle \SI{70}{\km\per\hour}, Test vehicle \SI{80}{\km\per\hour} \\ Spray vehicle \SI{100}{\km\per\hour}},
extra x tick style={
	major tick length=\pgfkeysvalueof{/pgfplots/minor tick length},
	tick label style={anchor=south},
	align=center,
},
extra y ticks={50},
extra y tick style={
	tick label style={red},
	grid style={red}
},
legend style={
	at={(0.865,\dimexpr-\pgfkeysvalueof{/pgfplots/minor tick length}+4.8cm\relax)},
	anchor=north,
	legend columns=8,
	font=\Large
},
]
\addplot[red,fill=red!30] table [percentage series=1] {\data};
\addplot[blue,fill=blue!30] table [percentage series=2] {\data};
 \legend{\strut Day,\strut Night}

\end{axis}
\end{tikzpicture}}
\caption{Blockage times of the camera sensor for the dynamic overtaking maneuvers. The blockage time is determined by the subjective assessment of the RGB images.}
\label{fig:camera_blockage}
\vspace*{-1mm}
\end{figure}
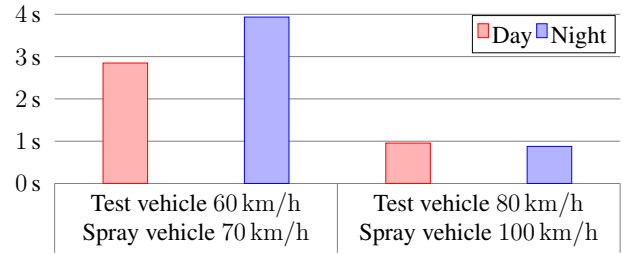

\begin{figure}[t]
\centering
 \vspace*{-2mm}
\resizebox{0.95\columnwidth}{!}{
  \begin{tikzpicture}
   \begin{axis}[
    xlabel={Speed [\unit{km/h}]},
    ylabel={Spray Length [\unit{m}]},
    grid=major,
    xmin=58,
    xmax=102,
    ymin=8,
    ymax=26,    
    legend style={
      cells={anchor=west},
      legend pos=north west,
      font=\footnotesize
    },
	axis line style={line width=1pt},
    legend entries={Day, Night},
    legend columns=2,
    legend pos= south east,
    width=\columnwidth,
    height=0.5\columnwidth,
    label style={font=\footnotesize,align=center},
    tick label style={font=\scriptsize},
    y label style={yshift=-1mm},
    x label style={yshift=0.5mm},
]

\addlegendimage{red!80, very thick}
\addlegendimage{blue!80, very thick}
\addplot+ 	[
thick,
red!80,
mark=o,
error bars/.cd,
y dir = both,
y explicit,
error bar style={line width=1pt}]
table[x index=0,y index=1, y error index=2, col sep=space]{fig/spray_length/spray_length.txt};

\addplot+ 	[
thick,
blue!80,
mark=o,
error bars/.cd,
y dir = both,
y explicit,
error bar style={line width=1pt},]
table[x index=0,y index=3,, y error index=4, col sep=space]{fig/spray_length/spray_length.txt};

  \end{axis}
 \end{tikzpicture}}
 \vspace*{-2mm}
 \caption{Length of the spray tail for different vehicle speeds.}
 \label{fig:spray_length}
 \vspace*{-6mm}
\end{figure}
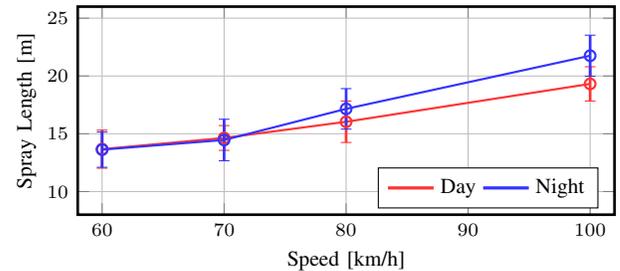

\begin{figure}[t]
\vspace*{0.7mm}
\centering
\resizebox{\columnwidth}{!}{
\begin{tikzpicture}
\draw[-,densely dashed](1.1,-0.4) -- (1.1,3.7);
\node[anchor=west,inner sep=0pt, text width=15mm, align=center] at (-0.35,3.2) {\footnotesize Cutting in \\[-1.5mm] \footnotesize process};
\draw[-,densely dashed](3.05,-0.4) -- (3.05,3.7);
\draw[white, pattern=north west lines, pattern color=high_contrast]  (1.1,-0.4)  -- (3.05,-0.4)  -- (3.05,3.7)-- (1.1,3.7) --cycle;
\node[anchor=west,inner sep=0pt, align=center] at (1.2,3.2) {\footnotesize Sensor setup in \\[-1.5mm] \footnotesize spray corridor \\[-1.5mm] \footnotesize ($\approx$ \SI{3}{\second})};
\node[anchor=west,inner sep=0pt, align=center] at (3.9,3.2) {\footnotesize Sensor setup out of \\[-1.5mm] \footnotesize spray corridor};
\begin{axis}[
ybar,
bar width=7pt,
legend style={at={(0.5,-0.2)},
	anchor=north,legend columns=-1, font=\scriptsize},
symbolic x coords={4-8\,m,8-12\,m,12-16\,m,16-20\,m,20-24\,m,24-28\,m,28-32\,m},
xtick={4-8\,m,8-12\,m,12-16\,m,16-20\,m,20-24\,m,24-28\,m,28-32\,m},
nodes near coords,
nodes near coords align={vertical},
every node near coord/.append style={rotate=90, anchor=west, font=\scriptsize},
y filter/.expression={y==0 ? nan : y},
unbounded coords=discard,
width=\columnwidth, height=4cm,
axis x line=bottom,
axis y line=left,
ymajorgrids=true,
ymax = 110,
ymin= 0,
enlarge x limits=0.1,
ylabel={Vehicle Recall [\%]},
label style={font=\footnotesize,align=center},
tick label style={font=\scriptsize}
]

\addplot[red,fill=red!30] coordinates {(4-8\,m,100) (8-12\,m,100)
	(12-16\,m,100) (16-20\,m,100) (20-24\,m,100)
	(24-28\,m,100) (28-32\,m,100)};
\addplot[blue,fill=blue!30] coordinates {(4-8\,m,50) (8-12\,m,0.0)
	(12-16\,m,6) (16-20\,m,41) (20-24\,m,77)
	(24-28\,m,88) (28-32\,m,100)};

\legend{Day,Night}
\end{axis}
\end{tikzpicture}}
\caption{Evaluation of the 2D object detector for different distances during overtaking in spray using the follwing speeds: Test vehicle: \SI{60}{\km\per\hour}, spray vehicle: \SI{70}{\km\per\hour}}
\label{fig:2d_detection_dynamic}
\vspace*{-2mm}
\end{figure}
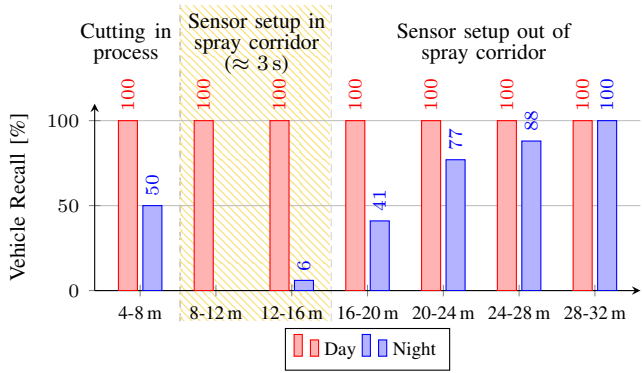
\begin{figure}[t]
\centering
\begin{subfigure}[t]{\columnwidth}
\centering
\subcaption{\small Recall rate of the spray vehicle ahead.}
\vspace*{-1mm}
\label{fig:3d_detection_dynamic_vehicle}
\begin{tikzpicture}
\draw[-,densely dashed](1.1,-0.4) -- (1.1,3.7);
\node[anchor=west,inner sep=0pt, text width=15mm, align=center] at (-0.35,3.2) {\footnotesize Cutting in \\[-1.5mm] \footnotesize process};
\draw[-,densely dashed](3.05,-0.4) -- (3.05,3.7);
\draw[white, pattern=north west lines, pattern color=high_contrast]  (1.1,-0.4)  -- (3.05,-0.4)  -- (3.05,3.7)-- (1.1,3.7) --cycle;
\node[anchor=west,inner sep=0pt, align=center] at (1.2,3.2) {\footnotesize Sensor setup in \\[-1.5mm] \footnotesize spray corridor \\[-1.5mm] \footnotesize ($\approx$ \SI{3}{\second})};
\node[anchor=west,inner sep=0pt, align=center] at (3.9,3.2) {\footnotesize Sensor setup out of \\[-1.5mm] \footnotesize spray corridor};
\begin{axis}[
ybar,
bar width=7pt,
legend style={at={(0.5,-0.2)},
	anchor=north,legend columns=-1, font=\scriptsize},
symbolic x coords={4-8\,m, 8-12\,m,12-16\,m,16-20\,m,20-24\,m,24-28\,m,28-32\,m},
xtick={4-8\,m, 8-12\,m,12-16\,m,16-20\,m,20-24\,m,24-28\,m,28-32\,m},
nodes near coords,
nodes near coords align={vertical},
every node near coord/.append style={rotate=90, anchor=west, font=\scriptsize},
unbounded coords=discard,
width=\columnwidth, height=4cm,
axis x line=bottom,
axis y line=left,
ymajorgrids=true,
ymax = 110,
ymin= 0,
enlarge x limits=0.1,
ylabel={Vehicle Recall [\%]},
label style={font=\footnotesize,align=center},
tick label style={font=\scriptsize}
]

\addplot[violet,fill=violet!30] coordinates {(4-8\,m,100) (8-12\,m,96)
	(12-16\,m,100) (16-20\,m,96) (20-24\,m,100)
	(24-28\,m,93) (28-32\,m, 100)};
\addplot[brown,fill=brown!30] coordinates {(4-8\,m,100) (8-12\,m,100)
	(12-16\,m,73) (16-20\,m,18) (20-24\,m,69)
	(24-28\,m,74) (28-32\,m,0.0)};

\end{axis}
\end{tikzpicture}
\end{subfigure}

\begin{subfigure}[t]{\columnwidth}
\vspace*{1mm}
\centering
\subcaption{\small False positive occurence rate of ghost objects in clutter.}
\vspace*{-1mm}
\label{fig:3d_detection_dynamic_ghost}
\begin{tikzpicture}
\draw[-,densely dashed](1.1,-0.4) -- (1.1,2.8);
\draw[-,densely dashed](3.05,-0.4) -- (3.05,2.8);
\draw[white, pattern=north west lines, pattern color=high_contrast]  (1.1,-0.4)  -- (3.05,-0.4)  -- (3.05,2.8)-- (1.1,2.8) --cycle;
\begin{axis}[
ybar,
bar width=7pt,
legend style={at={(0.5,-0.2)},
	anchor=north,legend columns=-1, font=\scriptsize},
symbolic x coords={4-8\,m,8-12\,m,12-16\,m,16-20\,m,20-24\,m,24-28\,m,28-32\,m},
xtick={4-8\,m,8-12\,m,12-16\,m,16-20\,m,20-24\,m,24-28\,m,28-32\,m},
nodes near coords,
nodes near coords align={vertical},
every node near coord/.append style={rotate=90, anchor=west, font=\scriptsize},
unbounded coords=discard,
width=\columnwidth, height=4cm,
axis x line=bottom,
axis y line=left,
ymajorgrids=true,
ymax = 110,
ymin= 0,
enlarge x limits=0.1,
ylabel={Ghost Recall [\%]},
label style={font=\footnotesize,align=center},
tick label style={font=\scriptsize}
]

\addplot[violet,fill=violet!30] coordinates {(4-8\,m,0) (8-12\,m,39)
	(12-16\,m,73) (16-20\,m,100) (20-24\,m,100)
	(24-28\,m,70) (28-32\,m, 39) };
\addplot[brown,fill=brown!30] coordinates {(4-8\,m,0) (8-12\,m,4)
	(12-16\,m,4) (16-20\,m,0) (20-24\,m,0)
	(24-28\,m,0) (28-32\,m,0) };

\legend{HDL64,VLP32}
\end{axis}
\end{tikzpicture}
\end{subfigure}
\caption{Evaluation of the 3D object detector for different distances during overtaking in spray using the follwing speeds: test vehicle: \SI{60}{\km\per\hour}, spray vehicle: \SI{70}{\km\per\hour}.}\label{fig:3d_detection_dynamic}
\vspace*{-6mm}
\end{figure}

For the quantitative evaluation of the static scenarios, ground truth annotations are determined by a human annotator in clear conditions. This allows calculating the \ac{AP} metric in each setting. Table~\ref{tab:2d_object_detection_static} provides the evaluation of the 2D object detector at day and night. Here, the impact of the cleaning system and the effect of active brake lights at night are studied.
To allow a fair evaluation of the detectors, images with visible wipers are filtered out.
The results show that without an activated cleaning system, the camera's view is totally blocked, and the detection algorithm is unable to recognize the spray vehicle ahead. 
The activation of the cleaning system leads to a significant increase in detection performance from zero to $70.9\%$ during day and $29.4\%$ at night.
With active brake lights at night, the cleaning system fails to improve the recognition. This indicates that the object detector is severely affected by the scattered lights from the bright brake lights, even if the windshield itself is clean.
Furthermore, the degradation of the \ac{AP} for higher \ac{IOU} thresholds indicates that the localization accuracy decreases due to spray.
This is most notable for the day setting with active cleaning where we note a drop from $70.9\%$ at $0.5$ \ac{IOU} to $15.2\%$ at $0.7$ \ac{IOU}.

For the quantitative evaluation of the laser scanners, measurements are recorded in coarse and fine spray conditions generated by the varying nozzles reported in Table~\ref{tab:nozzle-config} and shown in Table~\ref{tab:3d_object_detection_static}. Here, the average number of \ac{LIDAR} points detected by the Velodyne HDL64-S3D increases in spray conditions from 4993 to 5219 (5480 in fine spray) due to additional echos from the spray plume.
In contrast, the missing cleaning system of the Velodyne VLP32C leads to the blockage of the sensor and a decreased number of detected \ac{LIDAR} points, with an average of 2015 in clear conditions compared to 1001 in coarse spray and 128 in fine spray.
We assume that the larger and heavier droplets of the coarse spray help to keep the sensor clean by flowing straight off the sensor, while the lightweight droplets of the fine spray remain on the surface, blocking the sensor.
As a result, the object detector delivers much better results in coarse spray conditions than with fine spray for the Velodyne VLP32C.
The Velodyne HDL64-S3D on the other hand provides a constant performance for coarse and fine spray.
However, the significant degradation of the \ac{AP} for higher \ac{IOU} from $100\%$ to $77.1\%$ in coarse spray and from $96.2\%$ to $25\%$ in fine spray is remarkable and indicates localization problems caused by spray clutter. Cf. Table~\ref{tab:3d_object_detection_static} for all \ac{LIDAR} object detection results.

\begin{table}[t]
	\vspace*{1.7mm}
	\caption{Evaluation of the 2D object detector for the static scene. An active cleaning system increases the performance significantly, except at night and active brake lights. }
	\vspace*{-1mm}
	\centering
	\setlength{\tabcolsep}{1mm}
	\resizebox{\columnwidth}{!}{
		\begin{tabular}{@{}llcccccc@{}}
			\toprule
			&& & & \multicolumn{4}{c}{\textbf{night}}  \\ \cmidrule(rl){5-8}
			
			&& \multicolumn{2}{c}{\textbf{day}} & \multicolumn{2}{c}{\textbf{no brake}} &\multicolumn{2}{c}{\textbf{brake lights}}  \\ \cmidrule(rl){3-4} \cmidrule(rl){5-6} \cmidrule(rl){7-8}
			
			&cleaning& off & on & off & on & off & on \\
			
			\midrule
			
			\parbox[t]{4mm}{\multirow{3}{*}[-1mm]{\rotatebox[origin=c]{90}{\textbf{spray}}}} & \multirow{1}{*}[-1mm]{visual}  &\raisebox{-.5\height}{\includegraphics[width=0.075\textwidth]{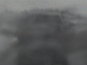}} & \raisebox{-.5\height}{\includegraphics[width=0.075\textwidth]{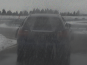}} & \raisebox{-.5\height}{\includegraphics[width=0.075\textwidth]{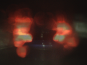}} & \raisebox{-.5\height}{\includegraphics[width=0.075\textwidth]{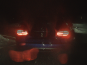}} & \raisebox{-.5\height}{\includegraphics[width=0.075\textwidth]{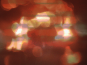}} & \raisebox{-.5\height}{\includegraphics[width=0.075\textwidth]{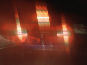}}  \\[4.5mm]

			& \small AP@0.5 & $3.08e^{-4}$ & 0.709 & $5.47e^{-3}$ & 0.294 & $1.64e^{-3}$ & $3.63e^{-2}$\\
			& \small AP@0.7 & 0.0 & 0.152 & $1.33e^{-4}$ & 0.218 & $1.71e^{-5}$ & $1.09e^{-3}$ \\

			\midrule
			
			\parbox[t]{4mm}{\multirow{3}{*}[-1mm]{\rotatebox[origin=c]{90}{\textbf{clear}}}} & \multirow{1}{*}[-1mm]{visual} & \multicolumn{2}{c}{\raisebox{-.5\height}{\includegraphics[width=0.075\textwidth]{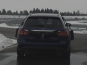}}}&
			\multicolumn{2}{c}{\raisebox{-.5\height}{\includegraphics[width=0.075\textwidth]{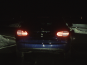}}}& \multicolumn{2}{c}{\raisebox{-.5\height}{\includegraphics[width=0.075\textwidth]{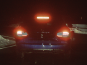}}}\\[4.5mm]
			
			& \small AP@0.5 & \multicolumn{2}{c}{1.0} & \multicolumn{2}{c}{1.0} & \multicolumn{2}{c}{0.939}\\
			
			& \small AP@0.7 & \multicolumn{2}{c}{1.0} & \multicolumn{2}{c}{1.0} & \multicolumn{2}{c}{0.862}\\

			\bottomrule
	\end{tabular}}
  	\label{tab:2d_object_detection_static}
\end{table}

\begin{table}[t]
	\vspace*{-2mm}
	\caption{Evaluation of the 3D object detector for the static scene and different spray conditions.}
	\vspace*{-1mm}
	\centering
	\setlength{\tabcolsep}{1mm}
	\resizebox{\columnwidth}{!}{
		\begin{tabular}{@{}cccccccc@{}}
			\toprule
			&& \multicolumn{3}{c}{\textbf{HDL64-S3D}} & \multicolumn{3}{c}{\textbf{VLP32C}}  \\ \cmidrule{3-5} \cmidrule(l){6-8}
			&\footnotesize{Nozzles}& \footnotesize{$\diameter$pts} & \footnotesize{AP@0.5} & \footnotesize{AP@0.7} & \footnotesize{$\diameter$pts} &  \footnotesize{AP@0.5} & \footnotesize{AP@0.7}\\
			\midrule
			\textbf{clear} & - &4993 & 1.0 & 1.0 & 2015 & 1.0 & 1.0\\
			\textbf{spray} & Type A&5219 & 1.0 & 0.771 & 1001 & 0.909 & 0.776\\
			\textbf{fine spray} & Type B &5480  & 0.962 & 0.25& 128 & 0.178 & 0.155\\
			\bottomrule
	\end{tabular}}
  	\label{tab:3d_object_detection_static}
  	\vspace*{-5mm}
\end{table}

\section{Conclusion and Outlook}
This article presents a spray machine for evaluating automotive sensors and object detection algorithms in reproducible spray conditions. Due to the construction of the spray machine, no modifications are required for the involved vehicle allowing to test with a variety of cars. 

In dynamic and static experiments, we demonstrated that the performance of both camera and \ac{LIDAR} sensors suffers from the impact of spray while the \ac{RADAR} system is not affected. The 2D object detection is constrained, especially at night, with active brake lights and a broken cleaning system. 
The spray causes a clutter plume in the \ac{LIDAR} point clouds affecting the appearance of ghost objects in downstream detection algorithms. The missing rotating housing of the Velodyne VLP32C leads to a thin water film in front of the optics resulting in an entire blockage. In static measurements, this caused a loss of 96 \% measurable points compared to clear conditions. 

Further, we envision that the spray setup and captured data enable the development of digital spray simulation techniques by altering clear captures or virtual data with real-world spray.

\section*{Acknowledgment}
The research leading to these results is funded by the Federal Ministry for Economic Affairs and Energy within the  project "VVM-Verification and Validation Methods for Automated Vehicles Level 4 and 5", a PEGASUS family project.  

\balance
\bibliographystyle{IEEEtran}
\bibliography{spray_test_track}
\end{document}